\documentclass[preprint,12pt]{elsarticle}

\usepackage{subcaption}
\usepackage{lineno,hyperref}
\usepackage{natbib}
\usepackage{geometry}
\usepackage{fleqn}
\usepackage{graphicx}
\usepackage{newtxtext,newtxmath}
\usepackage{hyperref}
\usepackage{booktabs}
\usepackage{bm}
\usepackage{amsmath}
\usepackage{enumitem}
\usepackage{threeparttable}
\usepackage{multirow}
\usepackage{esvect}
\usepackage{siunitx}
\usepackage{soul,xcolor}
\modulolinenumbers[5]

\graphicspath{{./figures/}}
\journal{Elsevier}

\usepackage{caption}
\usepackage[table]{xcolor}
\usepackage{siunitx}
\usepackage{etoolbox}
\appto\TPTnoteSettings{\footnotesize}

\begin{document}

\begin{frontmatter}

\title{\large{Assessment of Machine Learning--Based Critical Heat Flux Models in the CTF Subchannel Code for Square Rod Bundle Prediction}}

\author[NCSU]{Aidan Furlong\corref{mycorrespondingauthor}}
\cortext[mycorrespondingauthor]{Corresponding author}
\ead{ajfurlon@ncsu.edu}

\author[UWM]{Vinicius de Melo Monteiro}

\author[ORNL]{Robert Salko}

\author[UWM]{Juliana Pacheco Duarte}

\author[NCSU]{Xu Wu}

\address[NCSU]{Department of Nuclear Engineering, North Carolina State University,    \\ 
	Burlington Engineering Laboratories, 2500 Stinson Drive, Raleigh, NC 27695 \\}

\address[ORNL]{Nuclear Energy and Fuel Cycle Division, Oak Ridge National Laboratory, \\ 1 Bethel Valley Road, Oak Ridge, TN 37830}

\address[UWM]{Nuclear Engineering \& Engineering Physics Department, University of Wisconsin--Madison, \\ Engineering Research Building, 1500 Engineering Drive, Madison, WI 53711}

\begin{abstract}
% STATUS: DRAFTED
The prediction of critical heat flux (CHF), a key safety-related quantity in nuclear thermal hydraulics, remains an important challenge due to its direct relationship with fuel performance and reactor safety. Recent studies have demonstrated that relative to traditional empirical correlations and lookup tables (LUTs), machine learning (ML) methods can substantially improve CHF prediction accuracy. Most ML-based CHF models, however, have been developed and evaluated using tube databases, leaving their applicability to reactor-relevant rod bundle geometries largely unexplored.

This study evaluates ML-based CHF models deployed within the CTF subchannel code using the Electric Power Research Institute (EPRI) rod bundle CHF database. Both pure and hybrid residual correction models are considered in local and semilocal formulations. The tube-trained ML CHF models generally transferred favorably to rod bundle applications and outperformed traditional CHF methods across most geometries and operating conditions. The local hybrid LUT model produced the strongest overall performance, and the semilocal pure ML model remained highly competitive. Comparison against the Bowring correlation, W-3 correlation, and 2006 Groeneveld LUT demonstrated that substantial improvements in rod bundle CHF prediction are possible even when models are trained exclusively on tube data. These findings provide one of the first large-scale assessments of ML-based CHF models in square rod bundles within a production-level subchannel analysis environment and support their broader application in reactor thermal hydraulic analysis.
\end{abstract}


\end{frontmatter}

%\newpage
%%%%%%%%%%%%%%%%%%%%%%%%%%%%%%%%%%
\section{Introduction}
%%%%%%%%%%%%%%%%%%%%%%%%%%%%%%%%%%
% STATUS: DRAFTED

Critical heat flux (CHF), which marks the point at which heat transfer rapidly deteriorates due to unfavorable boiling mechanisms such as departure from nucleate boiling (DNB) or dryout, is a key performance quantity in boiling water systems. In nuclear reactors, which are typically operated under heat flux--controlled conditions, CHF can result in rapid increases in cladding temperature and a corresponding reduction in fuel safety margins. CHF is a safety-related parameter in nuclear reactors, and many operating reactors are ultimately limited by CHF, as it affects their operational flexibility and fuel utilization.

The prediction of CHF has been studied extensively for more than half a century. Historically these efforts have relied on experimental measurements followed by the development of empirical correlations and lookup tables (LUTs). Much of the available CHF database originates from uniformly heated tube experiments, which led to the development of widely used methods such as the Biasi correlation~\cite{biasi1967studies}, the Bowring correlation~\cite{bowring1972simple}, and the 2006 Groeneveld LUT~\cite{groeneveld20072006}. While these approaches remain widely used, their predictive accuracy is often inconsistent across operating conditions and geometries.

A fundamental challenge is that the majority of CHF experiments have been conducted in tubes, while reactor fuel assemblies are composed of rod bundles. Unlike isolated tubes, rod bundles experience additional physical phenomena, including turbulent mixing, crossflow, spacer grid effects, and nonuniform radial power distributions~\cite{todreas2021nuclear}. As such, prediction methods developed from tube data are not guaranteed to perform similarly in bundle environments. While bundle-specific correlations such as the W-3 correlation~\cite{tong1967heat} have been developed, the predictive performance of these traditional CHF methods in rod bundles remains an area of active interest.

In recent years, machine learning (ML) has emerged as an alternative data-driven approach for CHF prediction. This development has been accelerated by the public release of the US Nuclear Regulatory Commission (NRC) CHF database, which contains nearly 25{,}000 tube CHF measurements and was originally used in the construction of the 2006 Groeneveld LUT. Numerous studies have reported substantial improvements in predictive accuracy using ML techniques such as tree-based ML models, deep neural networks (DNNs), ensemble methods, and hybrid physics-informed approaches~\cite{qi2025machine,li2025prediction}. Among these approaches, hybrid residual learning models have received particular attention because they combine an established CHF methodology with an ML correction term, potentially improving both predictive accuracy and model interpretability~\cite{zhao2020prediction,khalid2023comparison,yang2024data,mao2024uncertainty,furlong2025physics}.

Despite these advances, most ML-based CHF studies remain focused on tube geometries. A limited number of investigations have considered rod bundles. Marcinkiewicz et al.~\cite{marcinkiewicz2022recurrent} employed recurrent neural networks for CHF prediction in a single $5\times5$ rod bundle geometry using local thermal hydraulic conditions generated by COBRA-FLX. Rohatgi et al.~\cite{rohatgi2022machine} explored the generation of synthetic bundle data using generative adversarial networks, and Khalid et al.~\cite{khalid2026effect} and Yang et al.~\cite{yang2026self} investigated bundle-based ML models using derived feature representations and transformer architectures. While these studies demonstrate the potential of ML in bundle applications, they are generally limited to specific geometries, bundle datasets, or modeling frameworks.

At the same time, ML-based CHF models have begun transitioning from stand-alone research studies into production thermal hydraulic analysis tools. Recent work has demonstrated the deployment of both pure and hybrid ML-based CHF models within subchannel analysis codes such as CTF and COBRA-TF~\cite{furlong2026deployment,mao2026evaluation}. However, these deployed models were trained primarily using tube data and have largely been evaluated using tube-based validation datasets. Consequently, relatively little is known about how well tube-trained ML CHF models transfer to rod bundle environments representative of reactor fuel assemblies.

This study evaluates the performance of tube-trained ML-based CHF models deployed within the CTF subchannel code when applied to square rod bundle geometries from the Electric Power Research Institute (EPRI) CHF database~\cite{fighetti1982parametric}. Both pure ML and hybrid residual correction models are considered in local and semilocal formulations and compared against conventional CHF prediction methods. Investigation includes quantifying the degree of domain shift between the tube training data and rod bundle deployment environment, assessing the influence of turbulent mixing assumptions, and evaluating predictive performance across a diverse range of bundle geometries and operating conditions.

Together, these analyses establish the extent to which production ML-based CHF models trained exclusively on tube data can be applied to reactor-relevant rod bundle simulations without retraining. In doing so, they provide one of the first large-scale assessments of tube-trained ML-based CHF models in rod bundle applications within a production-level subchannel analysis environment.

The remainder of this article is structured as follows: Section~\ref{sec:background} introduces the EPRI database and general ML-based CHF modeling considerations; Section~\ref{sec:methods} describes tube-based ML model development, compares tube and bundle database characteristics, details CTF simulation setup, and notes turbulent mixing as an uncertainty source; Section~\ref{sec:results} presents broad performance results, followed by a breakdown of model performance with respect to bundle geometry, axial power profile, and local thermal hydraulic conditions; and Section~\ref{sec:conclusions} provides concluding remarks and potential avenues for continued investigation.

%%%%%%%%%%%%%%%%%%%%%%%%%%%%%%%%%%
\section{Background}
\label{sec:background}
%%%%%%%%%%%%%%%%%%%%%%%%%%%%%%%%%%

%%%%%%%%%%%%%%%%%%%%%%%%%%%%%%%%%%
\subsection{EPRI rod bundle CHF database}

This study uses the EPRI rod bundle CHF database, compiled at the Columbia University Heat Transfer Research Facility (HTRF) and reported by Fighetti and Reddy~\cite{fighetti1982parametric}. Published under EPRI sponsorship in 1982, it is one of the largest publicly available rod bundle CHF databases. It contains 11{,}077 CHF measurements from 235 test sections~\cite{fighetti1982parametric}, contributed by several major nuclear vendors and laboratories and covering a wide range of light-water reactor (LWR) fuel assembly designs. Because the data come from many independent programs, the database includes a broad variety of bundle features, such as guide tubes, water tubes, different spacer grid designs, and both square- and triangular-pitch lattices.

The database covers a wide range of geometries and operating conditions~\cite{fighetti1982parametric}. Bundle sizes range from $3\times3$ to $6\times6$ square-pitch arrays, along with 19-, 28-, and 37-rod triangular-pitch configurations. The heated lengths range from about 0.8 to 4.3~\si{\meter}, and the diameters of the rods range from 9.5 to 19.8~\si{\milli\meter}, with uniform and nonuniform axial power profiles. The operating conditions span pressures of 1--17~\si{\mega\pascal}, mass fluxes of 50--6000~\si{\kilo\gram\per\square\meter\per\second}, inlet subcoolings of 7--1140~\si{\kilo\joule\per\kilo\gram}, and average heat fluxes of 0.2--4~\si{\mega\watt\per\square\meter}. These conditions are broadly representative of commercial LWR fuel.

\begin{figure}[ht!]
  \centering
  \begin{subfigure}{0.48\textwidth}
    \centering
    \includegraphics[width=\linewidth]{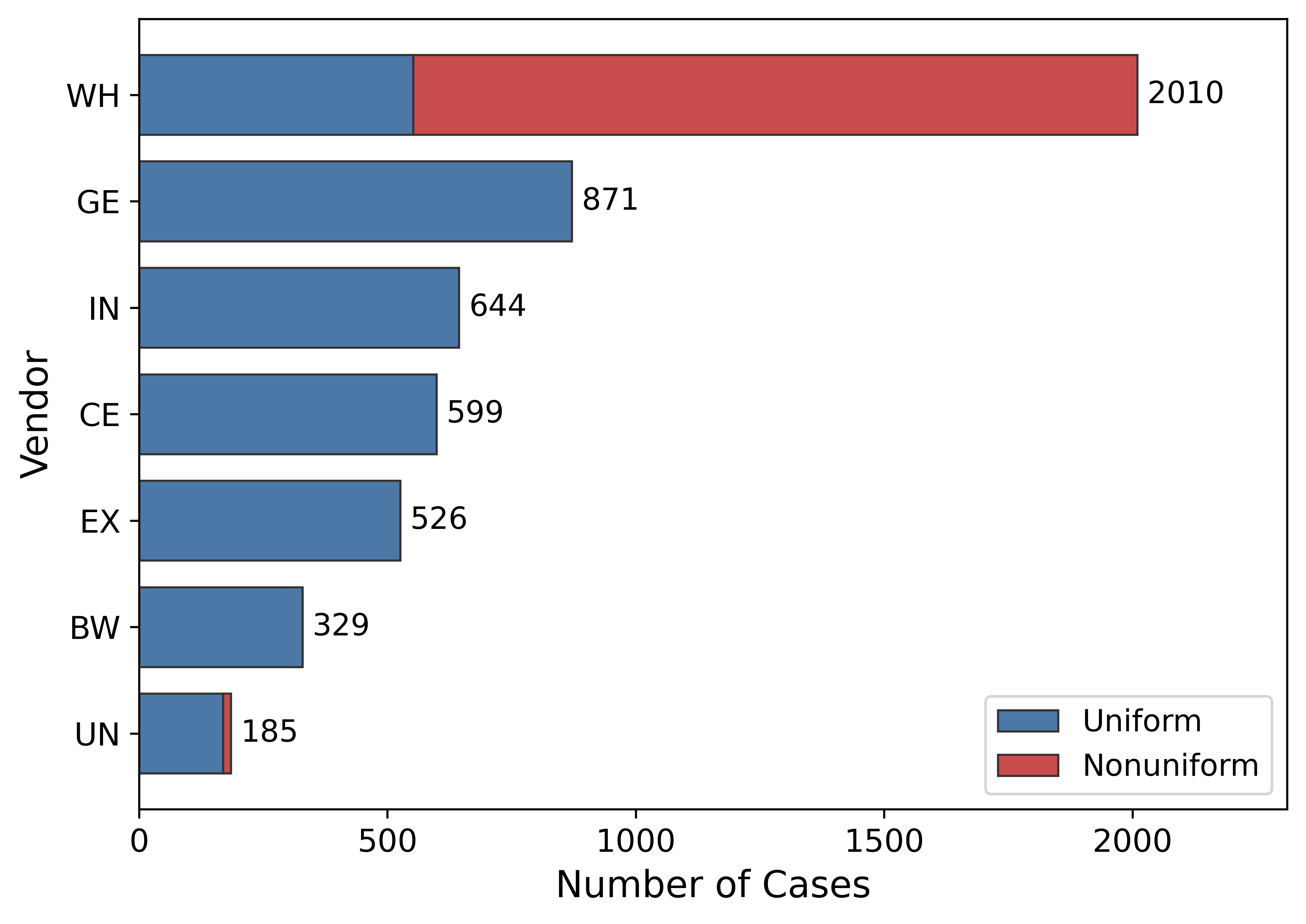}
    \caption{Axial power distribution}
  \end{subfigure}
  \hfill
  \begin{subfigure}{0.48\textwidth}
    \centering
    \includegraphics[width=\linewidth]{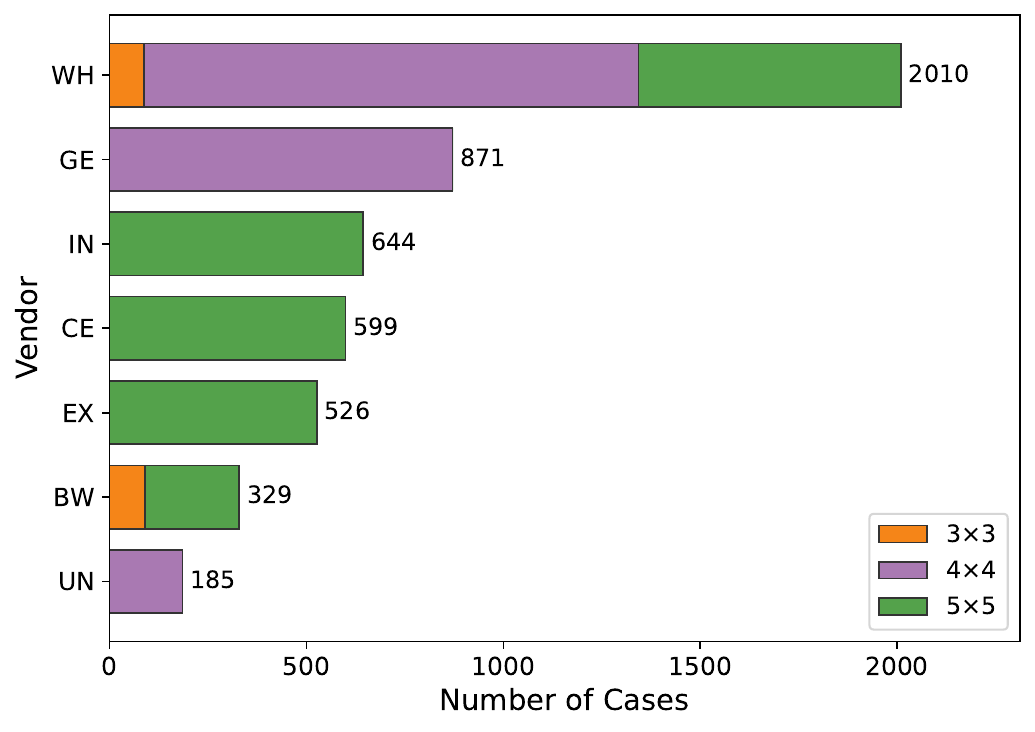}
    \caption{Bundle geometry type}
  \end{subfigure}
  \caption{Bundle characteristics across vendors in the EPRI square lattice subset. WH = Westinghouse Electric Co., GE = General Electric Co., IN = Idaho National Engineering Laboratories, CE = Combustion Engineering Inc., EX = Exxon Nuclear Co., BW = Babcock \& Wilcox Co., UN = United Nuclear Corp.}
  \label{fig:epri_subset}
\end{figure}

In this work, only the square lattice bundles $3\times3$, $4\times4$, and $5\times5$ are considered, under both uniform and nonuniform axial power profiles. Although other geometries present in the EPRI database (e.g., those with triangular pitch) are used in some operating reactor designs, square lattices remain the most common fuel design in the current LWR fleet. Figure~\ref{fig:epri_subset} shows the resulting subset by the contributing vendor, grouped by (a)~axial power profile and (b)~bundle geometry. Westinghouse provides the most cases, followed by General Electric, Idaho National Engineering Laboratories, Combustion Engineering, Exxon Nuclear, Babcock~\&~Wilcox, and United Nuclear. Most cases use uniform axial heating, but a substantial nonuniform subset is also retained, as shown in Figure~\ref{fig:epri_subset}.

%%%%%%%%%%%%%%%%%%%%%%%%%%%%%%%%%%
\subsection{Hybrid models and feature selection}
\label{subsec:background_chf_ml}
% STATUS: DRAFTED

The ML-based CHF models considered in this study were developed using the NRC public CHF database and represent the current ML-enabled CHF methodologies available within CTF. These models consist of both pure DNN-based ML and hybrid residual correction formulations and differ primarily in their input feature representations. The following discussion summarizes the hybrid modeling framework, underlying training database, and feature formulations relevant to the present validation study.

\begin{figure}[ht!]
    \centering
    \includegraphics[width=0.8\linewidth]{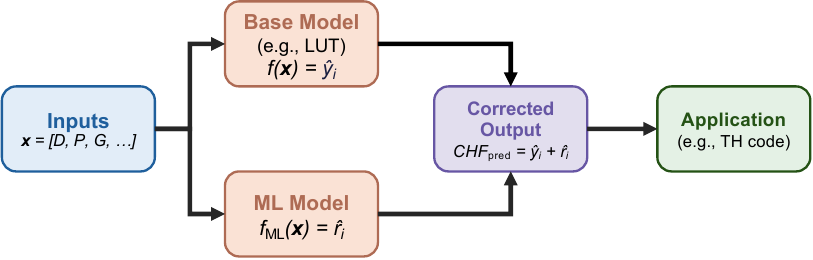}
    \caption{Deployment configuration of a hybrid residual learning ML model.}
    \label{fig:hybrid_model_workflow}
\end{figure}

Residual-learning \textit{hybrid} models seek to incorporate domain knowledge to improve predictive performance and reduce model variance~\cite{zhao2020prediction}. These models consist of a low-fidelity base model, such as an empirical correlation or LUT, coupled with an ML component, as illustrated in Figure~\ref{fig:hybrid_model_workflow}. Instead of directly predicting the target quantity, such as CHF, the ML component learns the discrepancy between the base prediction and the experimental value. This predicted residual is then used to correct the base model's output.

This bias correction framework can simplify the relationship that the ML component must learn since a portion of the underlying relationship is already represented by the base model. The degree of this simplification depends on the quality of the chosen base model, but in many cases, the residual is substantially easier to learn than the full CHF response. As a result of this design, hybrid models can exhibit lower variance and a more physically interpretable prediction process compared to equivalent ML models trained to predict CHF directly.

Much of the recent work in ML-based CHF modeling has followed the 2019 release of the NRC public CHF database~\cite{groeneveld2019critical}, originally used to construct the 2006 Groeneveld LUT. This 24,579-point database is composed of experimental test series from 59 references spanning more than 60 years, all of which consider uniformly heated tubes with water upflow and CHF occurring at the outlet. This dataset consists of seven measured/derived variables: diameter ($D$), tube heated length ($L_{\mathrm{h}}$), pressure ($P$), mass flux ($G$), enthalpy subcooling at the inlet ($\Delta h_{\mathrm{sub,in}}$), temperature at the inlet ($T_{\mathrm{in}}$), and the outlet equilibrium quality where CHF is detected ($x_{\mathrm{e}}$).

While the NRC database provides a large experimental foundation for model development, the choice of input features ultimately determines what physical relationships can be learned and where those models can be applied. In studies utilizing the NRC public database, input feature combinations generally fall into three categories: inlet condition--based, semilocal, and fully local formulations.

The inlet condition--based input feature combination depends entirely on a description of the tube inlet and is typically parameterized as $CHF(D, L_{\mathrm{h}}, P, G, \Delta h_{\mathrm{sub,in}} \ \mathrm{or} \ T_{\mathrm{in}})$. This formulation provides important upstream information that can help describe the CHF phenomenon when compared to an outlet-based parameterization. However, it learns relationships under the implicit assumptions of uniform heating, no mixing, no crossflow, and no other effects that would otherwise decouple the inlet from outlet conditions within an isolated subchannel. This substantially limits its applicability in environments such as rod bundle subchannels, which inherently exhibit these more complex behaviors.

The second formulation, which was primarily used in the OECD/NEA Phase I CHF ML Benchmark \cite{lecorre2023benchmark, lecorre2025oecd}, is a semilocal feature set $CHF(D,L_{\mathrm{h}},P,G,x_{\mathrm{e}})$. Instead of using an inlet condition as before, it uses the outlet equilibrium quality. Dependence on the outlet condition is favorable when considering model transferability, but the addition of heated length as an input introduces a weak dependency on the inlet. While heated length may serve as a surrogate for channel history, it does not uniquely determine the thermodynamic path leading to CHF. Consequently, multiple experiments may possess nearly identical $D, L_{\mathrm{h}}, P, G, x_{\mathrm{e}}$ values while exhibiting different CHF values due to differences in inlet conditions that are not observed by the model. This leaves only the correlation between heated length and CHF within the experimental database as the learned representation.

The fully local formulation does not depend on variables other than those found directly at the evaluation point, such as $CHF(D, P, G, x_{\mathrm{e}})$. This combination has the benefit of avoiding \textit{explicit} dependencies on experimental assumptions (such as uniform axial heating) or inlet conditions, as do the other formulations, but also lacks upstream information that is known to influence the CHF phenomenon. While this formulation has typically underperformed relative to the others on tube-based test data, it may provide the greatest transferability between geometries due to its reliance on purely local conditions.

Based on these considerations, only the fully local and semilocal formulations are considered in the present study. The inlet condition--based formulation is excluded due to its strong dependence on tube-specific inlet--outlet relationships that are not expected to remain valid within rod bundle subchannels. Evaluating the relative transferability of the local and semilocal formulations therefore constitutes a primary objective of this work.

%%%%%%%%%%%%%%%%%%%%%%%%%%%%%%%%%%
\section{Methods}
\label{sec:methods}
%%%%%%%%%%%%%%%%%%%%%%%%%%%%%%%%%%

%%%%%%%%%%%%%%%%%%%%%%%%%%%%%%%%%%
\subsection{Tube-based CHF models}
\label{sec:tube_models}
% STATUS: DRAFTED

Six ML-based CHF models are included in the current CTF distribution: pure ML, hybrid Bowring, and hybrid Groeneveld LUT in both local and semilocal formulations. All models were trained using data from the NRC tube public CHF database described in Section~\ref{subsec:background_chf_ml}.

Various references contained entries with negative inlet subcooling values, with unclear/differing methods in how these numbers were produced. Since traditional lookup of fluid properties using inlet temperature measurements is unreliable at near-saturation temperatures, these values were likely computed via other parameters and could be physically inconsistent. Therefore, these 258 entries were dropped, leaving 24,320 entries remaining. Since tubes are isolated subchannels, a simple heat balance can be easily applied to check whether input combinations produce local equilibrium qualities in agreement with those reported in the database. There can be several factors for disagreement---namely, measurement precision limitations, measurement bias, or simply the use of older fluid property LUTs such as IAPWS IF-67. There was limited disagreement between the computed and reported $x_{\mathrm{e}}$; for all model training, the heat balance--computed $x_{\mathrm{e}}$ values using IAPWS-IF97 fluid properties were implemented to maintain physical consistency.

The training data were then shuffled, partitioned into training/validation/testing datasets using a 90/5/5 split, and standardized (z-score normalization). This process produced a holdout dataset consisting of 1216 points reserved for final model evaluation prior to export. The primary model architecture is made up of five to seven densely connected hidden layers with varying widths, selected independently for each model type via a 1000-proposal hyperparameter optimization using the validation partition and the Optuna package~\cite{optuna_2019}. The final models were trained to a maximum of 500 epochs with early stopping to terminate learning once validation loss was satisfactorily converged. Learning rate decay and L2 regularization were also employed to aid in training and to reduce overfitting risks.

All model configurations were trained 20 times using different random initializations and training/validation samples to assess run-to-run variability. This permits the establishment of 95\% confidence intervals (CIs) on error metrics and helps determine whether the observed performance is representative of the model formulation or primarily the result of a favorable data split or initialization.

The resulting test set performance statistics are summarized in Table~\ref{tab:tube_model_performance_on_tubes}, where the ML-based approaches are additionally compared against the traditional CHF prediction methods available within CTF. Five metrics were used to characterize performance: the mean ($\upmu_{\mathrm{error}}$) and median ($\mathrm{Med}_{\mathrm{error}}$) absolute percentage error, the standard deviation ($\mathrm{Std}_{\mathrm{error}}$), and the fractions of cases with error above 10\% and 25\% ($F_{\epsilon>10\%}$ and $F_{\epsilon>25\%}$). The traditional CHF models in CTF are implemented using the direct substitution method (DSM), where CTF-computed values are used directly as inputs to the correlation or LUT~\cite{salko2023ctf}.

\begin{table}[ht!]
    \centering
    \caption{Test set performance on the NRC tube database. Values for ML-based methods denote the ensemble mean and associated 95\% CI across 20 independent training realizations. The W-3 correlation was excluded due to its limited range of validity.}
    \label{tab:tube_model_performance_on_tubes}
    \resizebox{\linewidth}{!}{
    \begin{tabular}{lccccc}
        \toprule
        \textbf{Method} &
        \boldmath $\upmu_{\textbf{error}}$ (\%) &
        $\textbf{Med}_{\textbf{error}}$ (\%) &
        $\textbf{Std}_{\textbf{error}}$ (\%) &
        \boldmath $F_{\epsilon>10\%}$ (\%) &
        \boldmath $F_{\epsilon>25\%}$ (\%) \\
        \midrule

        \textbf{Traditional methods}
        & & & & & \\
        
        \hspace{3mm} Bowring
        & 100.48
        & 32.51
        & 211.38
        & 80.08
        & 57.04 \\
        
        \hspace{3mm} LUT
        & 20.04
        & 11.86
        & 30.20
        & 55.56
        & 21.98 \\
        
        \textbf{ML-based, local}
        & & & & & \\
        
        \hspace{3mm} Pure ML
        & 13.77 $\pm$ 0.48
        & 7.61 $\pm$ 0.32
        & 24.53 $\pm$ 1.06
        & 40.45 $\pm$ 1.34
        & 13.27 $\pm$ 0.64 \\
        
        \hspace{3mm} Hybrid Bowring
        & 14.65 $\pm$ 0.63
        & 7.71 $\pm$ 0.35
        & 28.18 $\pm$ 1.62
        & 41.29 $\pm$ 1.30
        & 14.54 $\pm$ 0.90 \\
        
        \hspace{3mm} Hybrid LUT
        & 12.68 $\pm$ 0.32
        & 6.92 $\pm$ 0.23
        & 21.99 $\pm$ 0.63
        & 37.70 $\pm$ 0.96
        & 11.86 $\pm$ 0.59 \\
        
        \textbf{ML-based, semilocal}
        & & & & & \\
        
        \hspace{3mm} Pure ML
        & 8.84 $\pm$ 0.37
        & 5.43 $\pm$ 0.33
        & 12.40 $\pm$ 0.61
        & 28.44 $\pm$ 1.66
        & 6.27 $\pm$ 0.58 \\
        
        \hspace{3mm} Hybrid Bowring
        & 9.82 $\pm$ 0.36
        & 5.54 $\pm$ 0.26
        & 17.08 $\pm$ 1.30
        & 29.42 $\pm$ 1.20
        & 7.73 $\pm$ 0.40 \\
        
        \hspace{3mm} Hybrid LUT
        & 8.29 $\pm$ 0.13
        & 5.05 $\pm$ 0.14
        & 10.88 $\pm$ 0.39
        & 25.94 $\pm$ 0.64
        & 5.90 $\pm$ 0.29 \\
        
        \bottomrule
    \end{tabular}}
\end{table}

\begin{figure}[ht!]
    \centering
    \includegraphics[width=0.5\linewidth]{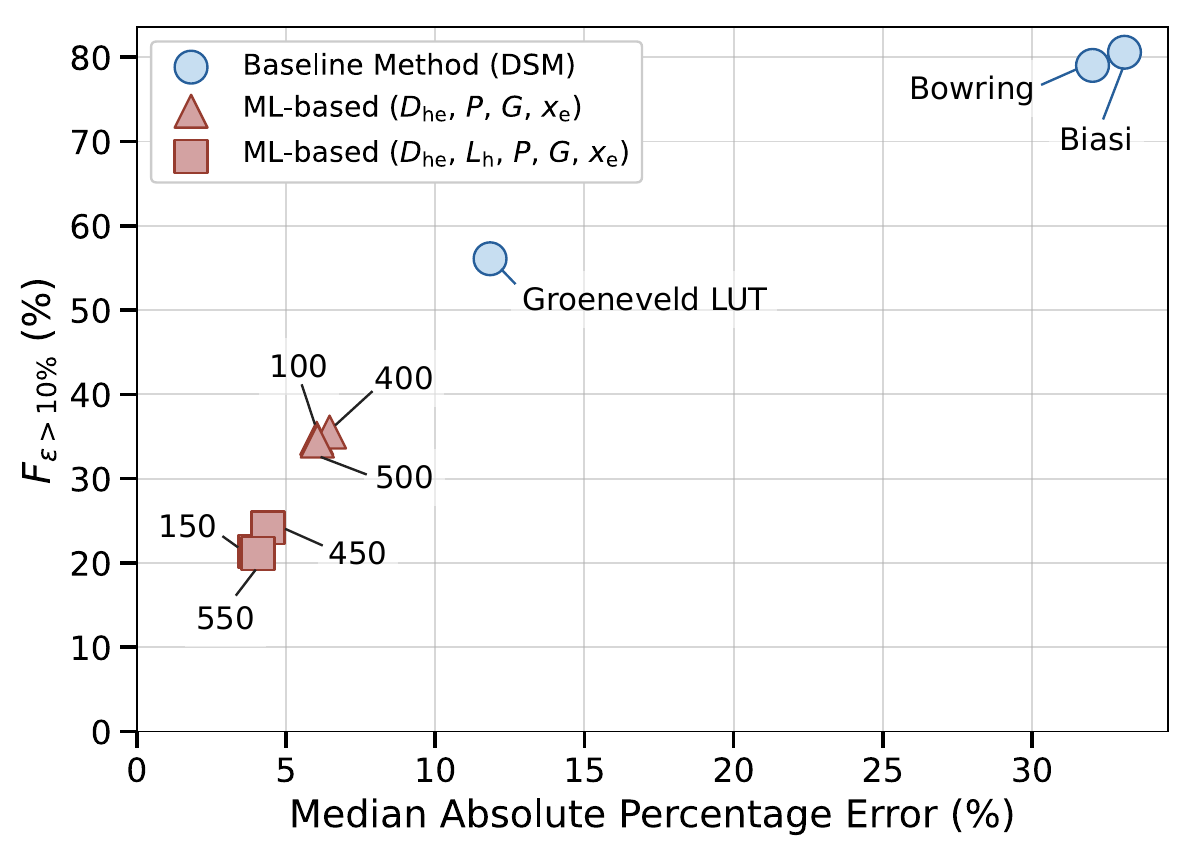}
    \caption{Tube-based CTF models performance on held-out NRC tube database as simulated in CTF. Models are labeled using their CTF identifier values: 1x0 = pure ML, 4x0 = hybrid Bowring, 5x0 = hybrid LUT. Hybrid Biasi model not included due to an open issue on CTF's Biasi correlation.}
    \label{fig:ctf_models_performance_tubes}
\end{figure}

Following training, a single realization from each model configuration was selected for deployment within CTF. The model parameters were then frozen, exported, and loaded directly into CTF using a previously developed open-source TensorFlow--Fortran bridge package~\cite{furlong2024fortrandeep,furlong2025native}. To characterize the specific models used throughout the remainder of this study, the deployed realizations were evaluated on the 1216-point held-out NRC tube test set using CTF simulations. Figure~\ref{fig:ctf_models_performance_tubes} summarizes the resulting median absolute percentage errors and fractions of predictions exceeding 10\% error and provides a direct comparison against the traditional CHF methods available within CTF.

%%%%%%%%%%%%%%%%%%%%%%%%%%%%%%%%%%
\subsection{CTF rod bundle models}

A CTF input deck was built for each retained square lattice test section from the geometry and operating data in the EPRI database~\cite{fighetti1982parametric}. Each bundle used a coolant-centered subchannel decomposition, where the control volume boundaries are straight lines joining adjacent rod centers. This is the standard control volume definition for DNB analysis in CTF~\cite{salko2023ctf} and resolves the lateral mass and momentum exchange between neighboring subchannels. The rods, subchannels, and gaps were defined from the bundle layout, and each spacer grid was represented by a local loss coefficient at its grid elevation. For every case, the deck imposed the measured inlet mass flow rate, inlet temperature, and outlet pressure, together with the experimental critical bundle power and its axial heat flux profile recorded at the onset of CHF.

The thermal hydraulic modeling configuration, summarized in Table~\ref{tab:ctf_config}, is based on the CTF setup previously validated for this database~\cite{monteiro2026}, providing a common and independently supported basis for the CHF method comparison. The same core models were retained, including the flow regime map, the Beus turbulent mixing and void drift formulation, the wall friction and nucleate boiling correlations, the thermophysical property tables, and the axial meshing strategy, with closure choices revisited where the present comparison required it. The cases were run with CTF v4.2-rc7, a two-fluid, three-field subchannel code that solves nine conservation equations for the vapor, continuous liquid, and entrained droplet fields over a vertical stack of control volumes coupled laterally through gaps~\cite{salko2023ctf}.

\begin{table}[ht!]
    \centering
    \caption{Summary of the CTF baseline modeling configuration used for the rod bundle simulations ~\cite{monteiro2026}.}
    \label{tab:ctf_config}
    \begin{tabular}{ll}
        \toprule
        \textbf{Category} & \textbf{Setting/value} \\
        \midrule
        CTF version & v4.2-rc7 \\
        Flow regime map & Legacy void-based map~\cite{salko2023ctf} \\
        Pressure matrix solver & Direct Gaussian elimination \\
        Thermophysical properties & IAPWS-IF97 steam tables~\cite{wagner2000iapws} \\
        Single-phase turbulent mixing$^{\dagger}$ & Rogers and Rosehart correlation~\cite{rogers1972mixing} \\
        Two-phase turbulent mixing & Beus multiplier $\Theta = 5.0$~\cite{beus1972two} \\
        Void drift & Equilibrium distribution coefficient $= 1.4$ \\
        Wall friction & $\lambda = 0.204\,\mathrm{Re}^{-0.2}$~\cite{mcadams1954heat} \\
        Nucleate boiling & Thom correlation~\cite{thom1965boiling} \\
        \bottomrule
    \end{tabular}
    
    \vspace{2pt}
    {\footnotesize $^{\dagger}$This choice differs from the configuration used in Monteiro et al.~\cite{monteiro2026} and is discussed in Section~\ref{sec:mixing}.}
\end{table}

To compare the CHF prediction methods on a common basis, nine methods were evaluated within CTF using the same set of rod bundle simulations. Applying every method to an identical thermal hydraulic solution isolates the effect of the CHF model itself so that any difference in predictive performance reflects the CHF formulation rather than differences in the underlying flow field. Three of the nine are traditional CHF methods: the W-3 correlation~\cite{tong1967heat}, the Bowring correlation~\cite{bowring1972simple}, and the 2006 Groeneveld LUT~\cite{groeneveld20072006}. The remaining six are the ML-based models introduced in Section~\ref{sec:tube_models}: the pure ML, hybrid Bowring, and hybrid LUT formulations, each evaluated in both local and semilocal formulations.

%%%%%%%%%%%%%%%%%%%%%%%%%%%%%%%%%%
\subsection{Influence of turbulent mixing}
\label{sec:mixing}
% STATUS: DRAFTED

\begin{figure}[ht!]
  \centering
  \begin{subfigure}{0.48\textwidth}
    \centering
    \includegraphics[width=\linewidth]{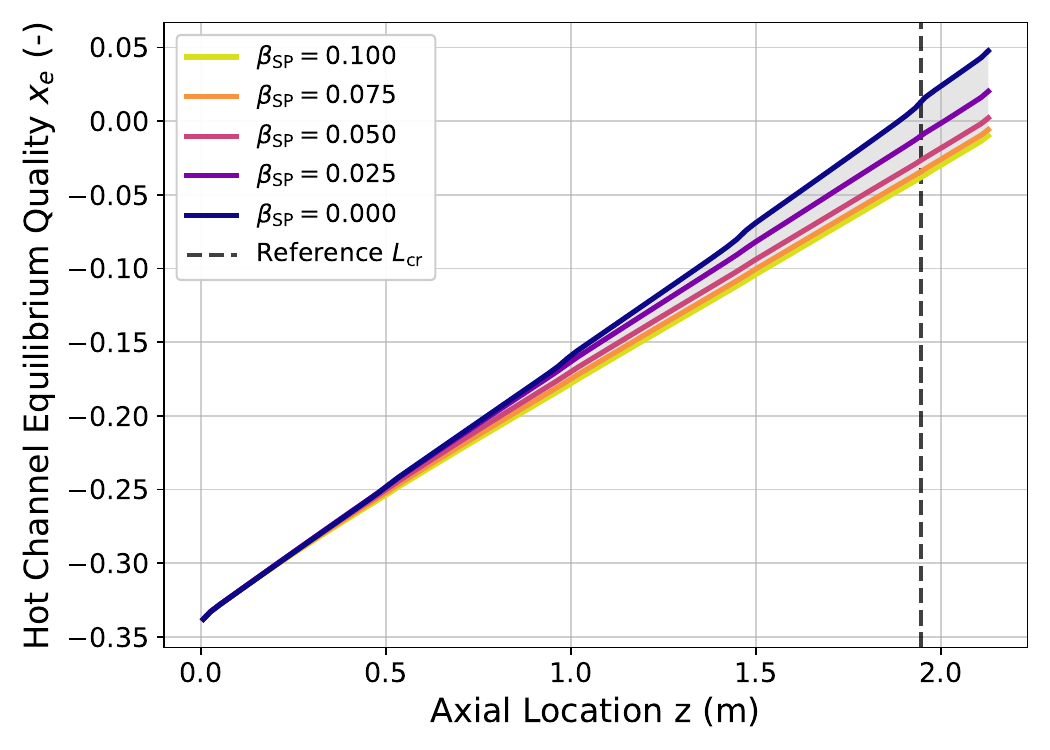}
    \caption{Equilibrium quality}
  \end{subfigure}
  \hfill
  \begin{subfigure}{0.48\textwidth}
    \centering
    \includegraphics[width=\linewidth]{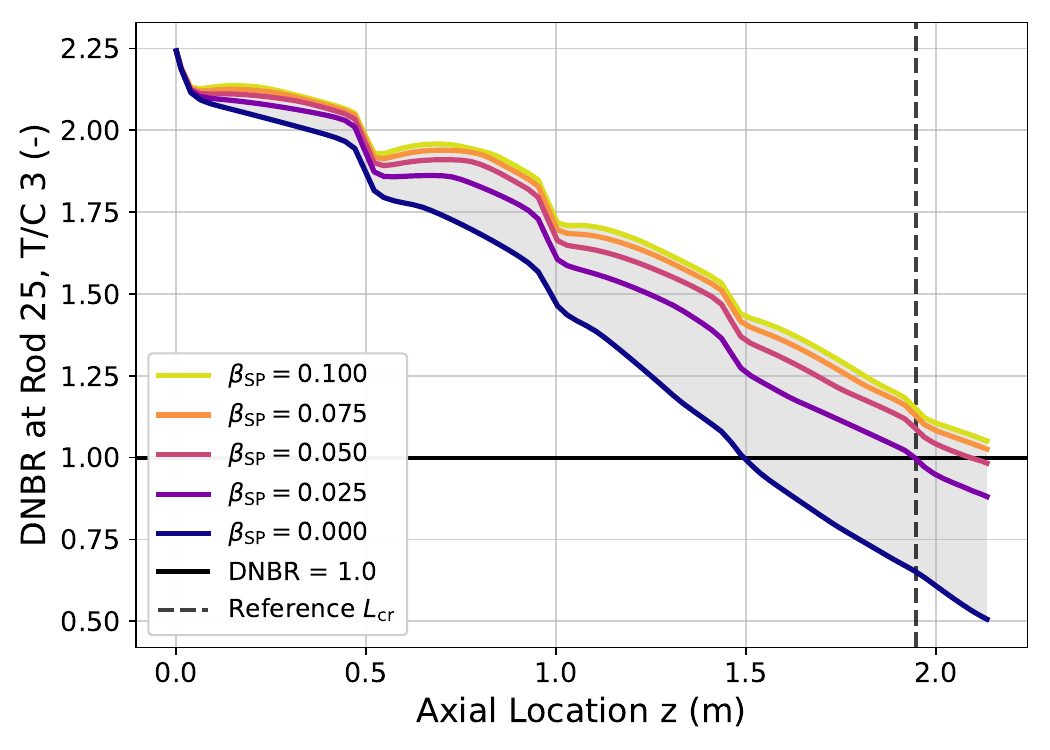}
    \caption{DNBR}
  \end{subfigure}
  \caption{Sensitivity of CTF simulation DNBR and channel equilibrium quality values to choice of $\beta_{\mathrm{SP}}$ in a representative CE $5\times5$ geometry using the Groeneveld LUT as the CHF model~\cite{furlong2026rodbundle}. The calibrated value for the shown testbed is 0.0044~\cite{sung2014application}.}
  \label{fig:beta_sensitivity}
\end{figure}

The choice of turbulent mixing treatment can significantly affect predicted local conditions and CHF model predictive performance. While other closure relations also contribute uncertainty, empirical investigations have consistently identified the turbulent mixing coefficient, $\beta$, as one of the most influential parameters in subchannel CHF analysis. The impact of this parameter is demonstrated in Figure~\ref{fig:beta_sensitivity}, which shows five otherwise identical CTF simulations performed for a reference rod bundle; the only parameter that was varied was the single-phase $\beta_{\mathrm{SP}}$ value. The critical location is indicated by the dashed line, where the departure from nucleate boiling ratio (DNBR) varies in excess of 45\% along with a significant spread in equilibrium quality values.

The EPRI report recommends a constant value of $\beta = 0.02$ for all bundle geometries. This value was obtained through a sensitivity study in which $\beta$ was varied and the resulting CHF correlation root mean square error was evaluated against experimental CHF measurements~\cite{fighetti1982parametric}. The study reported a large degree of sensitivity in local conditions to $\beta$, with variations approaching 60\% in subchannel enthalpy gain and 30\% in local mass flux. As a result of this optimization procedure, the recommended value is coupled to the CHF correlation used during calibration.

This distinction becomes important when comparing CHF methodologies. Calibration of a physical parameter against an imperfect modeling framework can produce an effective best fit value that compensates for model inadequacies rather than uniquely identifying the true physical parameter value~\cite{brynjarsdottir2014learning}. While a fixed value of $\beta = 0.02$ is reasonable for engineering analyses performed using a \textit{single} CHF methodology, its use is less straightforward when comparing \textit{multiple} CHF models whose rankings may depend on the treatment of turbulent mixing.

The objective of the present work is not to reproduce the historical EPRI CHF correlation workflow but rather to compare CHF methodologies under a common thermal hydraulic treatment that is not itself selected by optimizing CHF prediction error. The parameter $\beta$ is additionally known to vary with geometry and flow conditions~\cite{shen2018new}, with no universally accepted coefficient existing for square rod bundles~\cite{liu2020turbulent}. Although turbulent mixing coefficients are often calibrated using thermal mixing experiments and subchannel outlet temperature distributions~\cite{sung2014application}, such measurements are not available for the present database.

For these reasons, the Rogers and Rosehart~\cite{rogers1972mixing} correlation was adopted for the single-phase turbulent mixing coefficient together with the Beus two-phase multiplier~\cite{beus1972two}. It is important to note that this choice does not imply that the resulting $\beta$ values are uniquely correct. Rogers and Rosehart is, however, a classical experimentally derived turbulent mixing model that is used in several contemporary subchannel codes~\cite{chen2014strategies,salko2023ctf}, and it avoids directly determining the turbulent mixing coefficient through CHF prediction performance.

%%%%%%%%%%%%%%%%%%%%%%%%%%%%%%%%%%
\subsection{Comparison of tube and rod bundle operating domains}
\label{subsec:methods_domain_comparison}
% STATUS: DRAFTED

\begin{figure}[ht!]
    \centering
    \includegraphics[width=\linewidth]{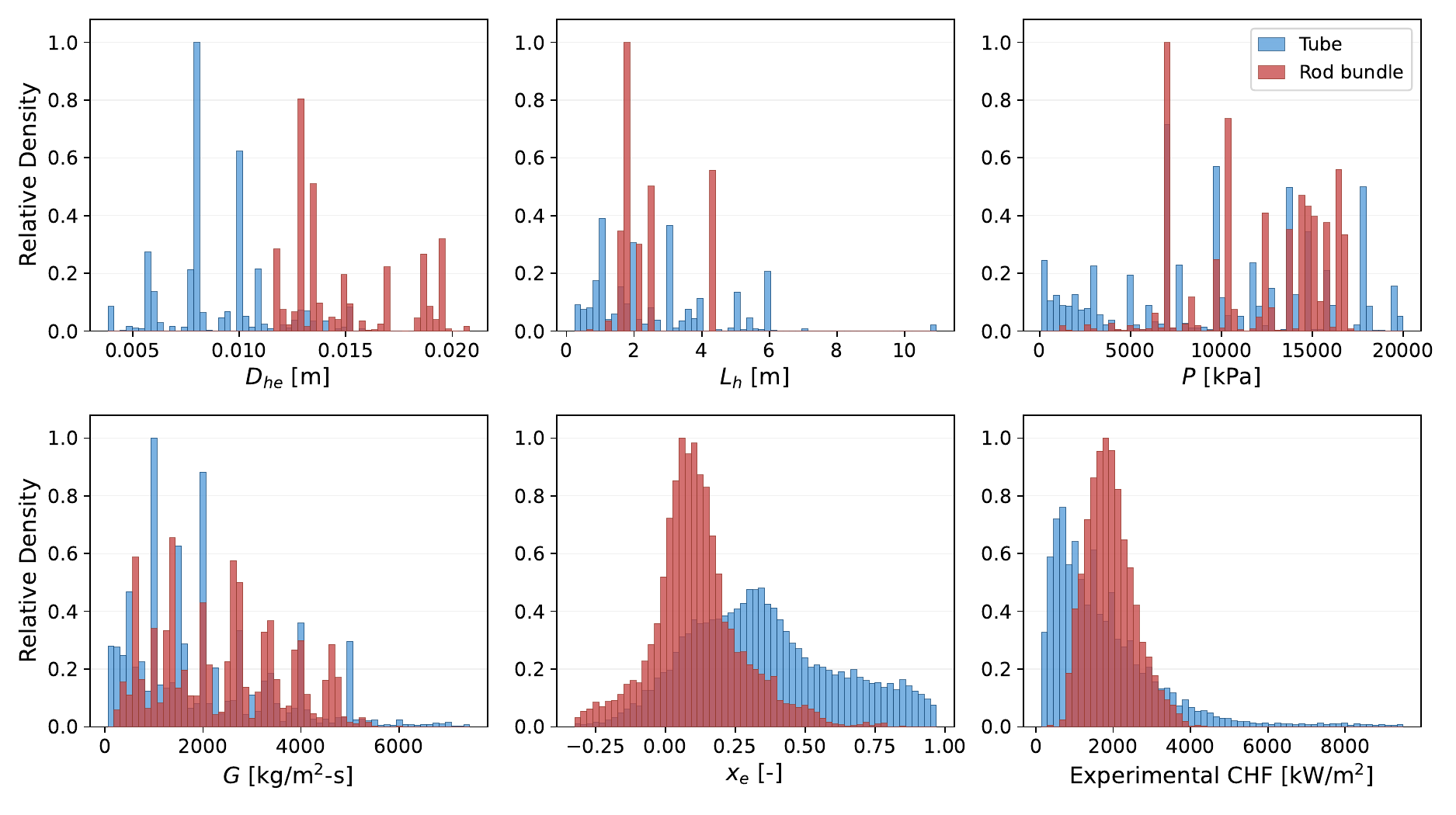}
    \caption{Marginal distributions of key input features in tube (training) and rod bundle (testing) datasets, measured with respect to the experimental CHF locations.}
    \label{fig:marginal_distributions_tube_bundle}
\end{figure}

Prior to evaluating predictive performance, it is useful to quantify the degree of overlap between the tube training domain and the rod bundle deployment domain. Figure~\ref{fig:marginal_distributions_tube_bundle} compares the marginal distributions of the ML model input variables for the tube training dataset and the rod bundle dataset used in this work. For the rod bundle cases, the input variables correspond to the local conditions predicted by CTF at the experimentally observed CHF location rather than inlet or bundle-averaged quantities. While substantial overlap exists between the two datasets, there are noticeable differences in several variables, particularly channel heated equivalent diameter ($D_{\mathrm{he}}$) and equilibrium quality. The pressure and mass flux distributions, however, exhibit considerable overlap with the tube database despite differences in their relative frequencies.

While marginal distributions provide a useful first assessment, they do not capture multivariate relationships between variables. To do this, several additional metrics were used to quantify the similarity between the rod bundle deployment domain and the tube training domain. The results are summarized in Table~\ref{tab:domain_overlap}. Min--max coverage represents the fraction of rod bundle states that reside within the tube training bounds (simple ranges) for all input variables simultaneously, while convex hull coverage provides a stricter measure of multivariate overlap. For both feature formulations, approximately 67.0\% of rod bundle states were contained within the tube training bounds. However, the convex hull coverage was substantially lower, indicating that many rod bundle states occupy combinations of variables that are not represented within the tube training dataset despite remaining within the individual variable ranges.

\begin{table}[ht!]
    \centering
    \caption{Comparison of rod bundle deployment domain states to the tube training domain for the local and semilocal feature formulations. Min--max coverage denotes the fraction of bundle states lying within the tube training bounds for all inputs simultaneously. Convex hull coverage denotes the fraction of bundle states residing within the multivariate tube training domain estimated using a Delaunay triangulation of a subset of tube observations. Nearest-neighbor distances were computed in tube-standardized feature space.}
    \label{tab:domain_overlap}
    \begin{tabular}{lcc}
        \toprule
        \textbf{Metric} & \textbf{Local} & \textbf{Semilocal} \\
        \midrule
        \textbf{Domain coverage} & & \\
        \hspace{3mm} Min--max coverage (\%) & 67.0 & 67.0 \\
        \hspace{3mm} Convex hull coverage (\%) & 48.7 & 35.6 \\
        \textbf{Bundle-to-training distances} & & \\
        % \hspace{3mm} Mean Distance & 0.746 & 0.883 \\
        \hspace{3mm} Median distance & 0.692 & 0.845 \\
        \hspace{3mm} 95th percentile distance & 2.125 & 2.234 \\
        \hspace{3mm} 99th percentile distance & 2.695 & 2.757 \\
        \textbf{Training data spacing} & & \\
        \hspace{3mm} Median distance & 0.024 & 0.033 \\
        \hspace{3mm} 95th percentile distance & 0.118 & 0.178 \\
        \hspace{3mm} 99th percentile distance & 0.219 & 0.328 \\
        \bottomrule
    \end{tabular}
\end{table}

Nearest-neighbor distances were also evaluated in a tube-standardized feature space, where the bundle-to-training distances quantify the separation between each rod bundle state and its nearest tube training sample, and the training data spacing metrics characterize the typical separation between tube training points. For both local and semilocal feature formulations, the rod bundle states were significantly farther from the tube training data than tube points were from each other. In the case of the local formulation, the median bundle-to-training distance for the local formulation was 0.692, compared to a median tube-to-tube spacing of only 0.024. Similar behavior was observed for the semilocal formulation, where the corresponding values were 0.845 and 0.033, respectively. These results indicate that although substantial overlap exists between the two datasets, many rod bundle operating conditions exist in relatively sparse regions of the tube training domain.

When assessing individual input features, the largest discrepancy was associated with heated equivalent diameter, where approximately 33.0\% of rod bundle states exceeded the maximum diameter represented in the tube training data. Analysis of the nearest-neighbor distances also identified heated equivalent diameter as the dominant contributor to the observed domain shift. For the local formulation, diameter accounted for 48.0\% of the average squared bundle-to-training distance and was the largest contributor for 57.5\% of rod bundle states. Similar behavior was observed for the semilocal formulation, where diameter accounted for 48.3\% of the average squared distance and was the dominant contributor for 57.7\% of rod bundle states.

This result is not necessarily surprising, as the physical interpretation of heated equivalent diameter fundamentally changes between the training and deployment environments. The heated equivalent diameter in the tube-based experimental design is identical to the physical tube diameter, while the rod bundle calculations employ a subchannel heated equivalent diameter that characterizes the local subchannel geometry. As such, the observed domain shift reflects both statistical differences in the variable distribution and a change in the underlying physical meaning of the variable itself.

The inclusion of heated length in the semilocal feature set reduced the measured overlap between the two domains. While the min--max coverage remained unchanged, the convex hull coverage decreased from 48.7\% for the local formulation to 35.6\% for the semilocal formulation. All bundle-to-training distance metrics also increased when heated length was included, indicating that heated length introduces an additional source of distributional shift despite all rod bundle states remaining within the tube heated length bounds. Overall, these results demonstrate that the local and semilocal tube-based models are routinely applied within portions of the rod bundle domain that are sparsely represented by the tube training data, with the dominant discrepancy arising from heated equivalent diameter.

%%%%%%%%%%%%%%%%%%%%%%%%%%%%%%%%%%
\section{Results}
\label{sec:results}
%%%%%%%%%%%%%%%%%%%%%%%%%%%%%%%%%%

A total of 4935 square-pitch rod bundle cases from the EPRI database were evaluated in CTF, of which 72 (about 1.5\%) failed to satisfy the strict convergence criteria. These failures were concentrated in the low mass flux regime, with about 85\% occurring below 500~kg\,m$^{-2}$\,s$^{-1}$, where the nonconvergence rate rose to roughly 25\% against near 1\% elsewhere. This is consistent with the known difficulty of subchannel solvers under weakly forced conditions: With little axial advection, buoyancy and two-phase volumetric expansion dominate, producing strongly nonlinear lateral coupling and possible counterflow that render the steady-state momentum and energy balances numerically stiff~\cite{zhu2023subchannel, salko2023ctf, salko2022ctf}. Low-flow conditions are also inherently prone to physical flow instabilities that further increase this difficulty~\cite{groeneveld20072006}. Because relaxing the tolerances would admit unacceptably large residual errors in the local fluid conditions on which the CHF assessment depends, these cases were excluded. Since the subchannel conservation equations are solved independently of the CHF correlation applied in post-processing, the same 72 cases failed for every method, so all nine methods are evaluated over the identical 4863 cases, and the reported differences reflect the CHF formulations themselves rather than a shifting case composition.

Predictive performance was quantified from the absolute relative error between the predicted and measured CHF, $\epsilon_i = |\mathrm{CHF_{pred}} - \mathrm{CHF_{meas}}| / \mathrm{CHF_{meas}}$. The measured critical bundle power was imposed as a fixed boundary condition, and CTF computed the local fluid state. For each run, the solution was searched over all rods, channels, surfaces, and axial nodes, the location of the minimum DNBR was identified as the limiting location~\cite{zhao2019prediction}, and $\mathrm{CHF_{pred}}$ was extracted there; $\mathrm{CHF_{meas}}$ is the experimental value reported for that case. Because the local conditions are computed by CTF at the imposed experimental power, $\epsilon_i$ reflects the combined error arising from both the CHF correlation and the CTF-predicted local state.

%%%%%%%%%%%%%%%%%%%%%%%%%%%%%%%%%%
\subsection{Overall rod bundle prediction performance}
% STATUS: DRAFTED

Aggregate error metrics are reported in Table~\ref{tab:tube_model_performance_on_bundles}. Both the W-3 correlation and the 2006 Groeneveld LUT perform similarly across most metrics, with median absolute error values close to 24\% and corresponding standard deviations close to 36\%. The baseline Bowring correlation is observed as having a significantly larger set of error metrics, with a median value of 36.89\% and 86.04\% of points above 10\% error.

\begin{table}[ht!]
    \centering
    \caption{Aggregate CHF predictive performance on the entire square rod bundle evaluation dataset, organized by CHF model type. Grey cells denote those with the most favorable error metrics.}
    \label{tab:tube_model_performance_on_bundles}
    \resizebox{\linewidth}{!}{
    \begin{tabular}{lccccc}
        \toprule
        \textbf{Method} &
        \boldmath $\upmu_{\textbf{error}}$ (\%) &
        $\textbf{Med}_{\textbf{error}}$ (\%) &
        $\textbf{Std}_{\textbf{error}}$ (\%) &
        \boldmath $F_{\epsilon>10\%}$ (\%) &
        \boldmath $F_{\epsilon>25\%}$ (\%) \\
        \midrule
        \textbf{Traditional methods}
        & & & & & \\
        \hspace{3mm} W-3
        & 31.69
        & 24.50
        & 36.34
        & 79.40
        & 48.96 \\
        
        \hspace{3mm} Bowring
        & 54.09
        & 36.89
        & 68.66
        & 86.04
        & 64.57 \\
        
        \hspace{3mm} LUT
        & 28.63
        & 23.47
        & 36.57
        & 80.24
        & 45.71 \\
        
        \textbf{ML-based, local}
        & & & & & \\
        
        \hspace{3mm} Pure ML
        & 25.86
        & 20.24
        & 30.65
        & 75.16
        & 38.70 \\
        
        \hspace{3mm} Hybrid Bowring
        & 39.07
        & 28.67
        & 36.98
        & 83.59
        & 56.71 \\
        
        \hspace{3mm} Hybrid LUT
        & 22.56
        & \cellcolor{gray!25} 15.46
        & 30.67
        & \cellcolor{gray!25} 67.74
        & \cellcolor{gray!25} 29.10 \\
        
        \textbf{ML-based, semilocal}
        & & & & & \\
        
        \hspace{3mm} Pure ML
        & \cellcolor{gray!25} 21.78
        & 18.33
        & \cellcolor{gray!25} 25.14
        & 73.41
        & 31.96 \\
        
        \hspace{3mm} Hybrid Bowring
        & 30.35
        & 20.55
        & 33.96
        & 73.80
        & 41.29 \\
        
        \hspace{3mm} Hybrid LUT
        & 27.33
        & 21.76
        & 30.03
        & 75.65
        & 43.24 \\
        
        \bottomrule
    \end{tabular}}
\end{table}

The ML-based models generally produced more favorable error metrics than the traditional methods. Performance, however, varied considerably between input feature formulations and hybrid model types. The strongest overall performance was obtained by the local hybrid LUT model, which produced the lowest median error (15.46\%), fraction of predictions above 10\% error (67.74\%), and fraction of predictions above 25\% error (29.10\%), while the mean error instead favored the semilocal pure ML model. Relative to the Groeneveld LUT, this corresponds to a reduction in median error of approximately 34\% and a reduction in $F_{\epsilon>25\%}$ of approximately 36\%. The semilocal pure ML model produced the lowest mean error (21.78\%) and the lowest error standard deviation (25.14\%) and remained highly competitive across all other metrics.

The use of hybrid models did not universally improve performance. While the local hybrid LUT model was the strongest-performing method overall, the semilocal hybrid LUT model was observed as having noticeably poorer performance than its pure ML counterpart. The local hybrid Bowring model similarly retained much of the error behavior associated with the underlying Bowring correlation and was the only ML-based model that did not consistently outperform the traditional methods. These results indicate that the effectiveness of residual learning is strongly dependent on both the selected base model and feature formulation.

\begin{figure}[ht!]
    \centering
    \includegraphics[width=0.5\linewidth]{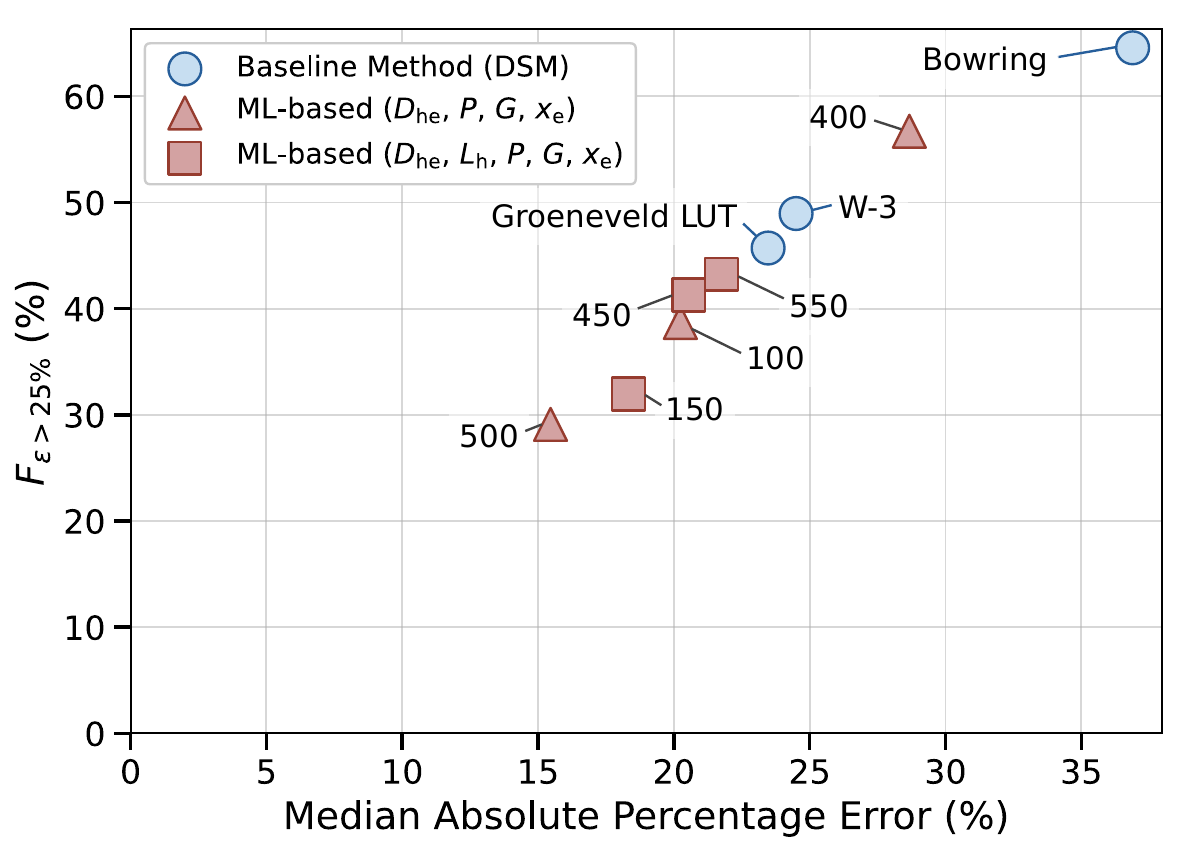}
    \caption{Tube-based CTF model performance on the EPRI square bundle database subset as simulated in CTF. Models are labeled using their CTF identifier values: 1x0 = pure ML, 4x0 = hybrid Bowring, 5x0 = hybrid LUT.}
    \label{fig:ctf_models_performance_card_bundles}
\end{figure}

To visualize model performance as was done in the case of tubes, Figure~\ref{fig:ctf_models_performance_card_bundles} plots each model using their global $\mathrm{Med}_{\mathrm{error}}$ and $F_{\epsilon > 25\%}$ values. With the exception of model 400 (local hybrid Bowring), all ML-based models show more favorable positions compared to the conventional comparators. The Bowring correlation, as noted earlier, reports a significantly larger error than other models, directly followed by the local hybrid Bowring model. No coherent trend is visible between local and semilocal ML model formulations.

\begin{figure}[ht!]
    \centering
    \includegraphics[width=\linewidth]{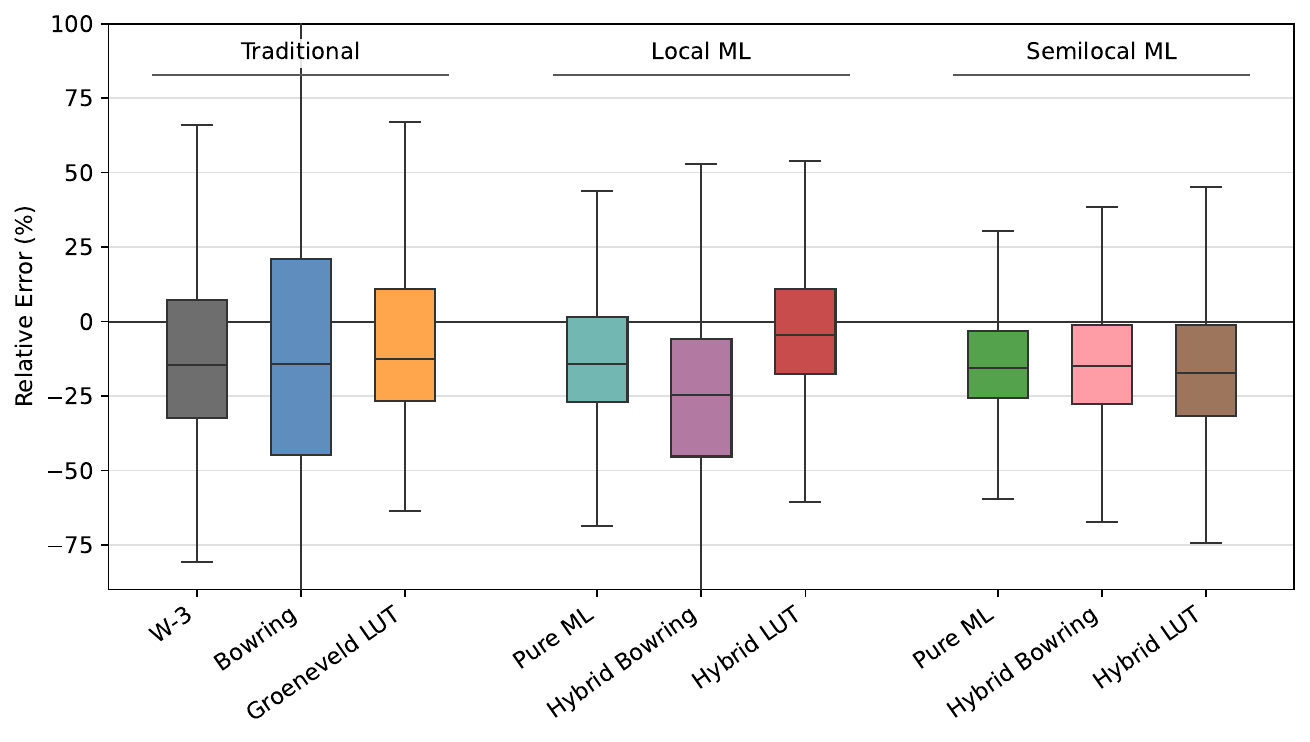}
    \caption{Visualization of the CHF models' signed relative error distributions.}
    \label{fig:signed_relative_error_aggregate}
\end{figure}

While analysis of absolute error is necessary to assess prediction magnitude, examination of signed relative error provides insight into systematic model bias. The distributions of relative error for each CHF model are shown as box plots in Figure~\ref{fig:signed_relative_error_aggregate}. Each distribution is characterized by its median (central line), interquartile range (IQR, box height), and nonoutlier data range (whiskers). All evaluated CHF models exhibit negative median relative errors, indicating a general tendency to underpredict CHF. Among the evaluated methods, the local hybrid LUT model reports the smallest bias, with a median closest to zero and the most balanced distribution of positive and negative error values.

With the exception of the local hybrid Bowring model, the ML-based approaches generally exhibit smaller IQRs and narrower overall error ranges than the traditional methods, indicating more consistent predictive behavior. The semilocal pure ML model produces the tightest overall distribution, suggesting the lowest prediction variance, while the local hybrid LUT model achieves the smallest systematic bias. Together, these observations are consistent with the aggregate error metrics previously noted, where both the local hybrid LUT model and the semilocal pure ML model consistently outperformed the other models.

%%%%%%%%%%%%%%%%%%%%%%%%%%%%%%%%%%
\subsection{Performance by test characteristics: bundle geometry}
% STATUS: DRAFTED

While aggregate error metrics provide a useful measure of overall performance, the EPRI database contains several bundle geometries that may exhibit different CHF prediction behavior. To attempt to separate these effects, the results were grouped according to bundle geometry and are summarized in Table~\ref{tab:bundle_geometry_performance}.

\begin{table}[ht!]
    \centering
    \caption{CHF predictive performance by bundle geometry. Entries are reported as $\text{Med}_{\text{error}}$ (\%) / $F_{\epsilon>25\%}$ (\%).}
    \label{tab:bundle_geometry_performance}
    \begin{tabular}{lccccc}
        \toprule
        \textbf{Method}
        & $\mathbf{3\times3}$
        & $\mathbf{4\times4}$
        & $\mathbf{5\times5}$ \\

        & $n=176$
        & $n=2291$
        & $n=2396$ \\
        \midrule

        \textbf{Traditional methods} \\

        \hspace{3mm} W-3
        & 29.05 / 56.82
        & 24.55 / 51.42
        & 22.08 / 46.04 \\
        
        \hspace{3mm} Bowring
        & 40.87 / 71.02
        & 38.28 / 68.79
        & 30.46 / 60.06 \\

        \hspace{3mm} LUT
        & 25.55 / 52.27
        & 25.42 / 53.56
        & 19.84 / 37.73 \\

        \textbf{ML-based, local} \\

        \hspace{3mm} Pure ML
        & 26.12 / 52.27
        & 19.71 / 39.07
        & 18.22 / 37.35 \\

        \hspace{3mm} Hybrid Bowring
        & 30.34 / 60.80
        & 34.01 / 61.59
        & 25.20 / 51.75 \\

        \hspace{3mm} Hybrid LUT
        & \cellcolor{gray!25} 18.19 / 34.66
        & \cellcolor{gray!25} 13.40 / 21.65
        & 16.11 / 35.81 \\

        \textbf{ML-based, semilocal} \\

        \hspace{3mm} Pure ML
        & 18.53 / 37.50
        & 20.02 / 36.58
        & \cellcolor{gray!25} 15.35 / 27.13 \\

        \hspace{3mm} Hybrid Bowring
        & 22.78 / 40.91
        & 17.75 / 41.86
        & 18.83 / 40.78 \\

        \hspace{3mm} Hybrid LUT
        & 21.17 / 36.93
        & 21.58 / 42.86
        & 20.67 / 44.07 \\

        \bottomrule
    \end{tabular}
\end{table}

The local hybrid LUT model produced the most favorable performance for both the $3\times3$ and $4\times4$ bundle geometries. The largest improvement was observed in the $4\times4$ bundles, where the median error decreased from 25.42\% for the Groeneveld LUT to 13.40\%, while $F_{\epsilon>25\%}$ decreased from 53.56\% to 21.65\%. Similar improvements were observed in the $3\times3$ bundles, where the local hybrid LUT model reduced the median error to 18.19\%.

The behavior of the $5\times5$ bundles differed somewhat from the smaller geometries. While the local hybrid LUT model remained highly competitive, the semilocal pure ML model produced the most favorable performance, with a median error of 15.35\% and only 27.13\% of predictions exceeding 25\% error. This represents a substantial improvement relative to the Groeneveld LUT, which reported a median error of 19.84\% and an $F_{\epsilon>25\%}$ value of 37.73\%.

Across all geometries, the Bowring correlation consistently produced the least favorable results among the traditional methods. Similar behavior was observed for the local hybrid Bowring model, indicating that residual learning was unable to fully compensate for deficiencies in the baseline correlation. The use of residual correction with the Groeneveld LUT, however, produced the strongest-performing model overall.

\begin{figure}[ht!]
    \centering
    \includegraphics[width=\linewidth]{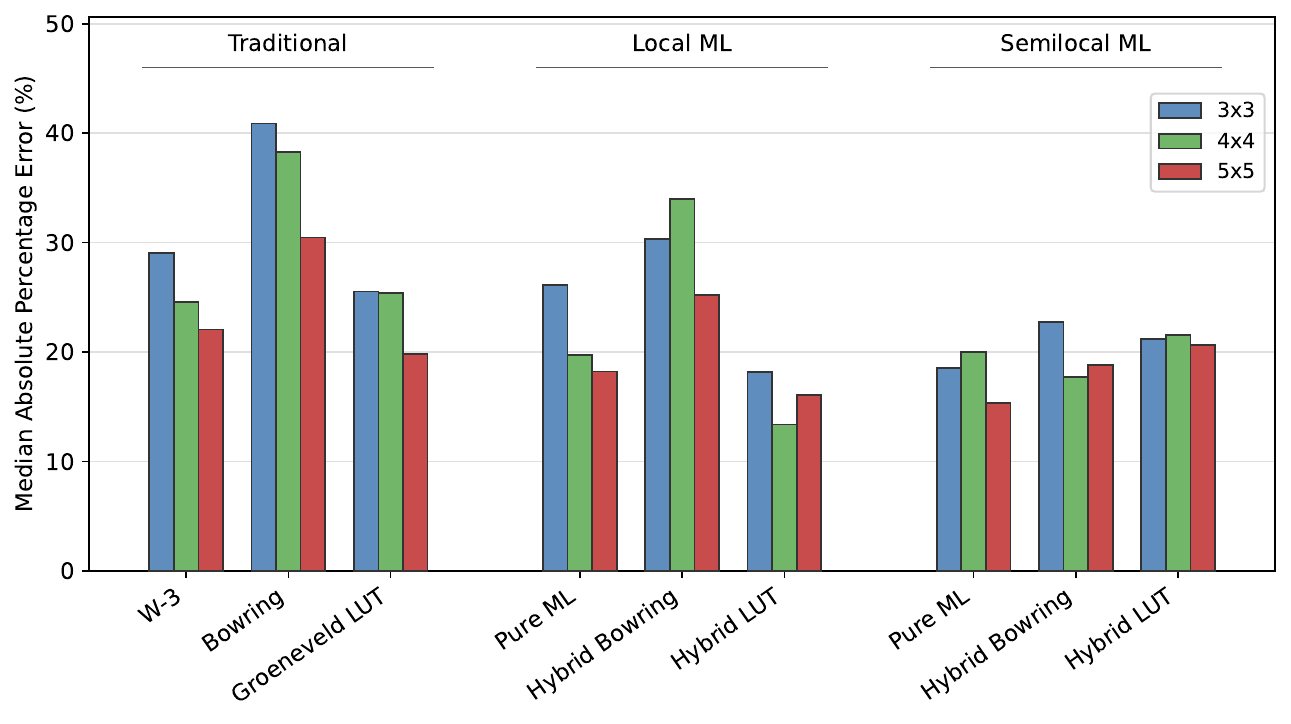}
    \caption{Visualization of the CHF models' performance by lattice size.}
    \label{fig:medape_by_bundle_geometry}
\end{figure}

Figure~\ref{fig:medape_by_bundle_geometry} shows the median absolute percentage error for each model across the three bundle geometries. No single model was uniformly dominant across all geometries, and no clear separation between local and semilocal formulations was observed. While the local hybrid LUT model was the strongest-performing model overall, the semilocal pure ML model produced the most favorable results for the largest bundle geometry.

%%%%%%%%%%%%%%%%%%%%%%%%%%%%%%%%%%
\subsection{Performance by test characteristics: axial power shape}

Another way to separate the test cases is by the axial power shape, which can be either uniform or nonuniform. Most of the EPRI square lattice experiments were performed using uniform power profiles (3396 cases), although a substantial number (1467 cases) employed nonuniform shapes that are more representative of operating fuel assemblies. The resulting performance metrics are summarized in Table~\ref{tab:bundle_profile_performance}.

\begin{table}[ht!]
    \centering
    \caption{CHF predictive performance by axial power shapes. Entries are reported as $\text{Med}_{\text{error}}$ (\%) / $F_{\epsilon>25\%}$ (\%). Two nonuniform column entries are highlighted, as they each have one of two metrics outperforming all other models.}
    \label{tab:bundle_profile_performance}
    \begin{tabular}{lcc}
        \toprule
        \textbf{Method}
        & \textbf{Uniform}
        & \textbf{Nonuniform} \\

        & $n=3396$
        & $n=1467$ \\
        \midrule

        \textbf{Traditional methods} \\

        \hspace{3mm} W-3
        & 21.40 / 45.64
        & 27.94 / 56.65 \\

        \hspace{3mm} Bowring
        & 33.07 / 64.69
        & 37.90 / 64.28 \\

        \hspace{3mm} LUT
        & 21.44 / 43.55
        & 25.00 / 50.72 \\

        \textbf{ML-based, local} \\

        \hspace{3mm} Pure ML
        & 19.81 / 41.46
        & 18.44 / 32.31 \\

        \hspace{3mm} Hybrid Bowring
        & 32.77 / 64.05
        & 20.96 / 39.74 \\

        \hspace{3mm} Hybrid LUT
        & \cellcolor{gray!25} 14.62 / 32.18
        & \cellcolor{gray!25} 14.91 / 21.95 \\

        \textbf{ML-based, semilocal} \\

        \hspace{3mm} Pure ML
        & 17.56 / 32.63
        & 18.41 / 30.40 \\

        \hspace{3mm} Hybrid Bowring
        & 20.49 / 48.35
        & \cellcolor{gray!25} 14.63 / 24.95 \\

        \hspace{3mm} Hybrid LUT
        & 23.06 / 47.94
        & 18.15 / 32.38 \\

        \bottomrule
    \end{tabular}
\end{table}

Most ML-based models outperformed the traditional CHF methods for both uniform and nonuniform power distributions. The strongest overall performance was again obtained by the local hybrid LUT model, which produced the lowest error metrics for both power profile categories with the exception of the nonuniform median error, where the semilocal hybrid Bowring model reported a slightly lower value (14.63\% versus 14.91\%). The local hybrid Bowring model also exhibited an interesting improvement under nonuniform conditions, even producing lower errors than the traditional CHF methods.

Previously, it was hypothesized that the local formulations could be better suited for rod bundle prediction due to the absence of an explicit heated length dependency. If this were the dominant effect, one might expect the local models to consistently outperform the semilocal models for nonuniform power profiles. This behavior was not observed. Instead, the semilocal models generally exhibited similar or lower errors under nonuniform conditions compared to the uniform cases. These results suggest that the inclusion of heated length does not inherently degrade model transferability under nonuniform power distributions.

\begin{figure}[ht!]
    \centering
    \includegraphics[width=\linewidth]{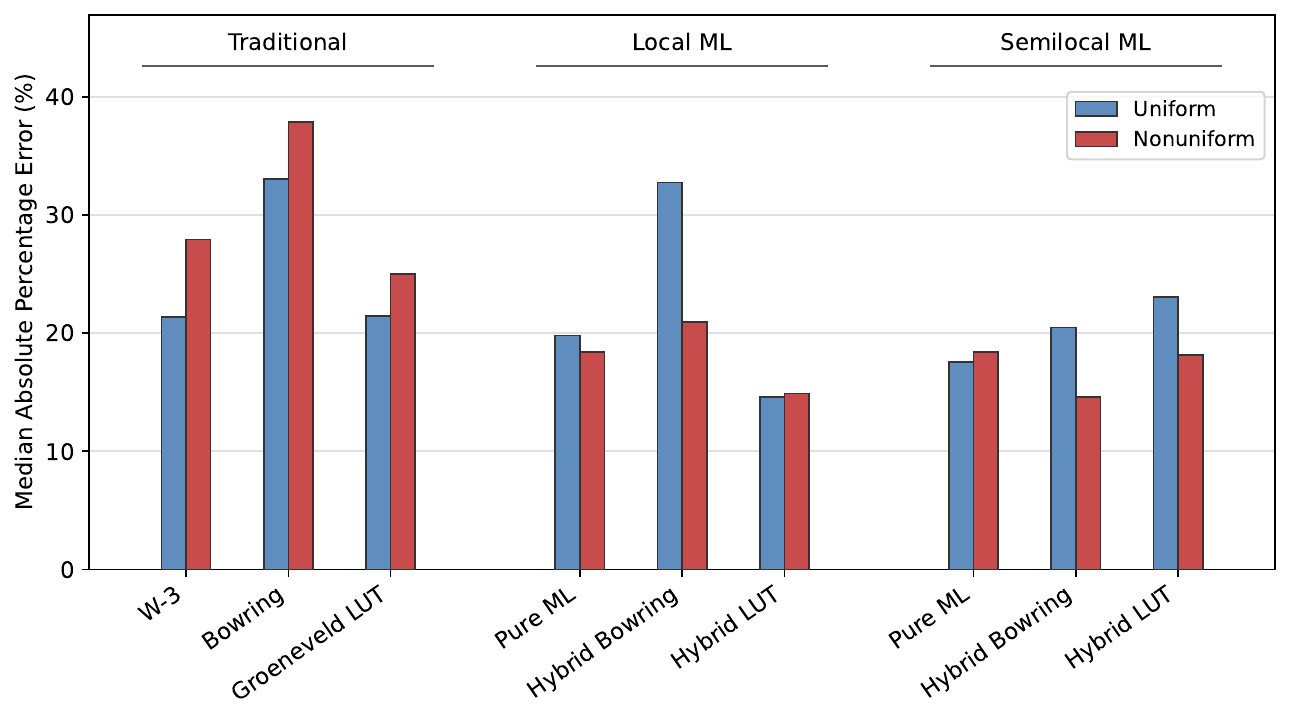}
    \caption{Visualization of the CHF models' performance by axial power profile.}
    \label{fig:medape_by_axial_profile}
\end{figure}

The median errors for each model are shown in Figure~\ref{fig:medape_by_axial_profile}. One immediately observable trend is that most ML-based models maintained similar or improved performance under nonuniform conditions, while the traditional CHF methods generally exhibited larger errors. Overall, the local hybrid LUT model demonstrated the most consistent performance between the two power profile categories and remained the strongest-performing model across nearly all evaluated metrics.

%%%%%%%%%%%%%%%%%%%%%%%%%%%%%%%%%%
\subsection{Performance by test characteristics: operating conditions}

\begin{figure}[ht!]
  \centering
  \begin{subfigure}{0.49\textwidth}
    \centering
    \includegraphics[width=\linewidth]{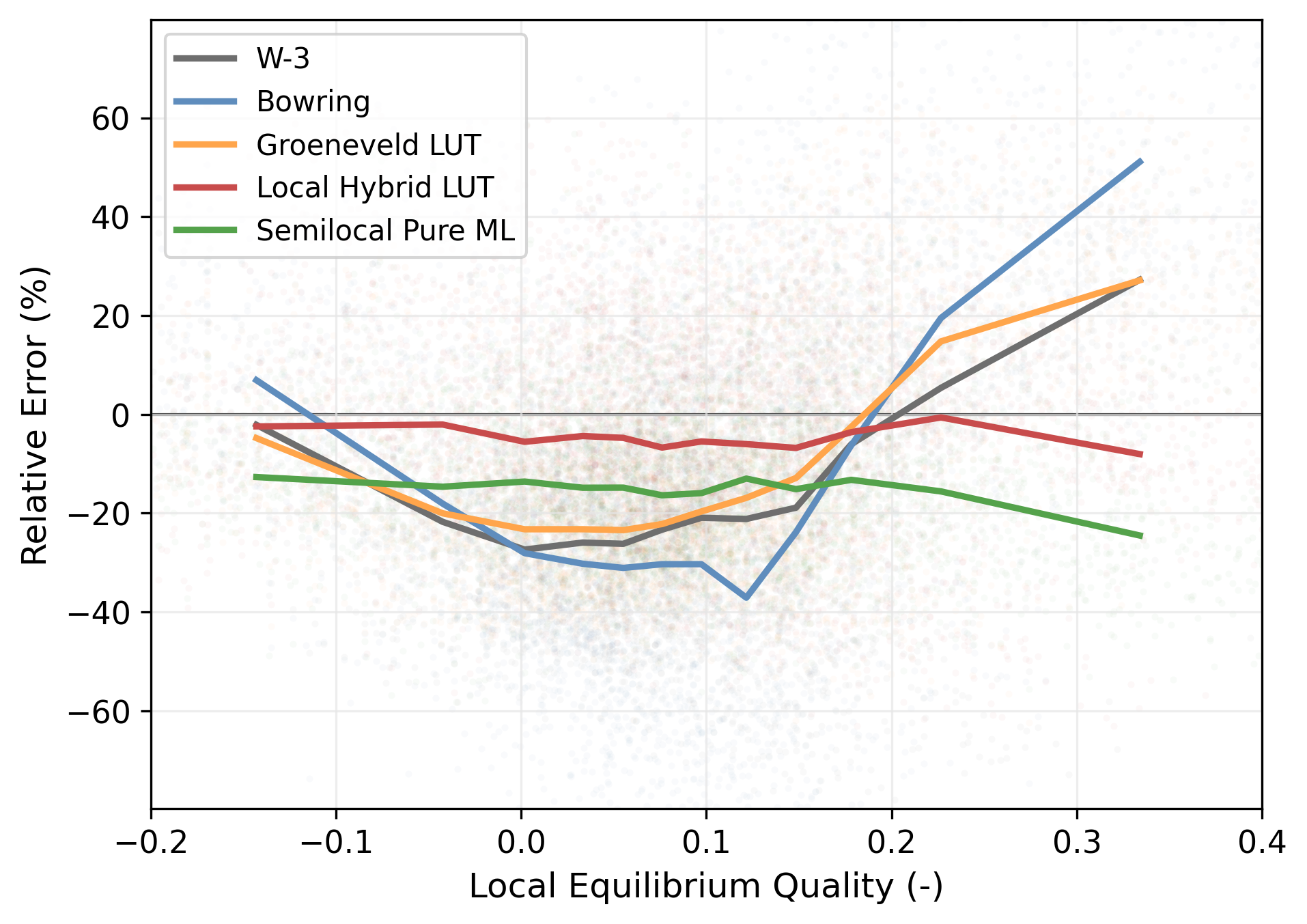}
    \caption{Equilibrium quality}
  \end{subfigure}
  \hfill
  \begin{subfigure}{0.49\textwidth}
    \centering
    \includegraphics[width=\linewidth]{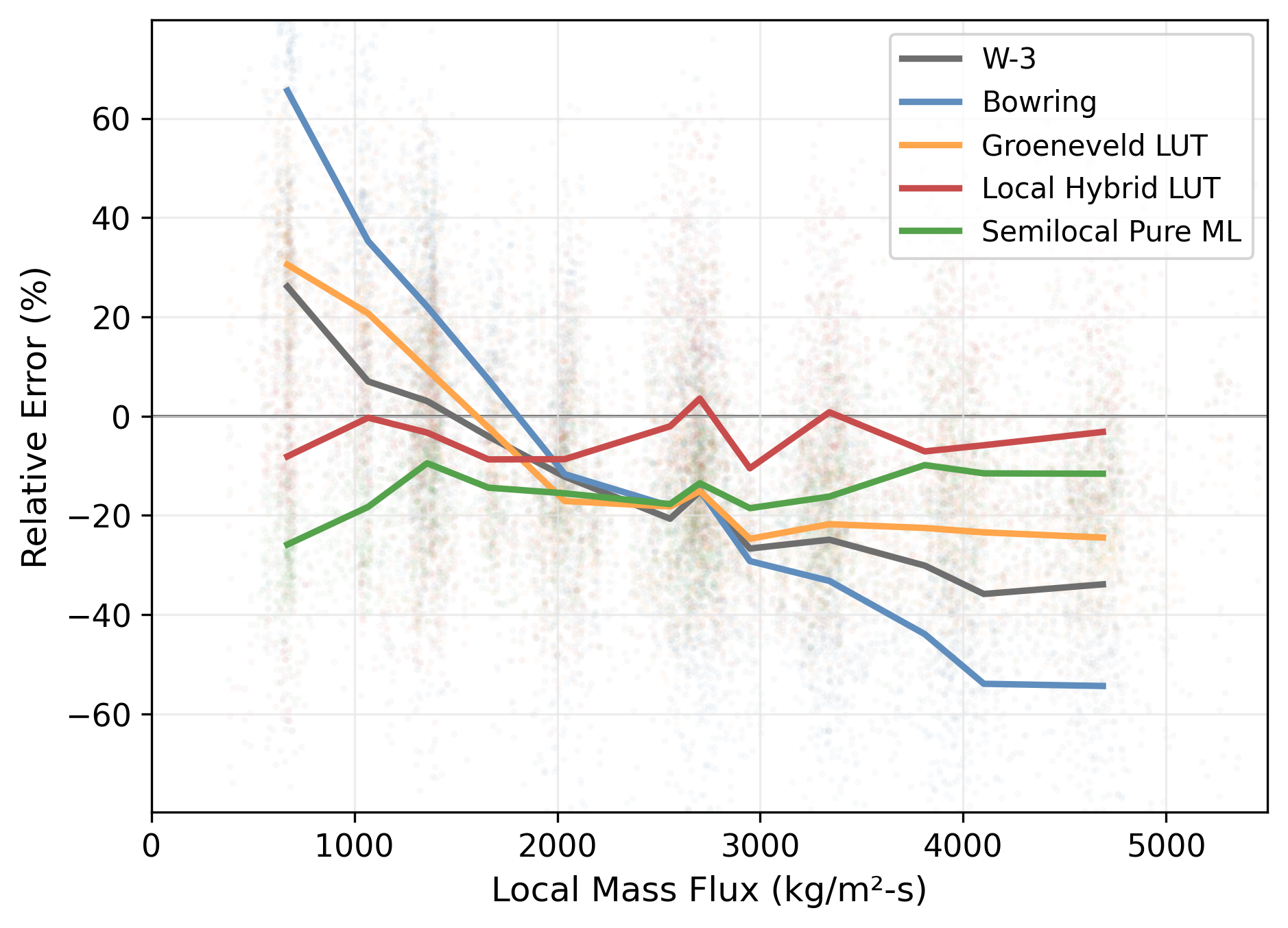}
    \caption{Mass flux}
  \end{subfigure}
  \begin{subfigure}{0.49\textwidth}
    \centering
    \includegraphics[width=\linewidth]{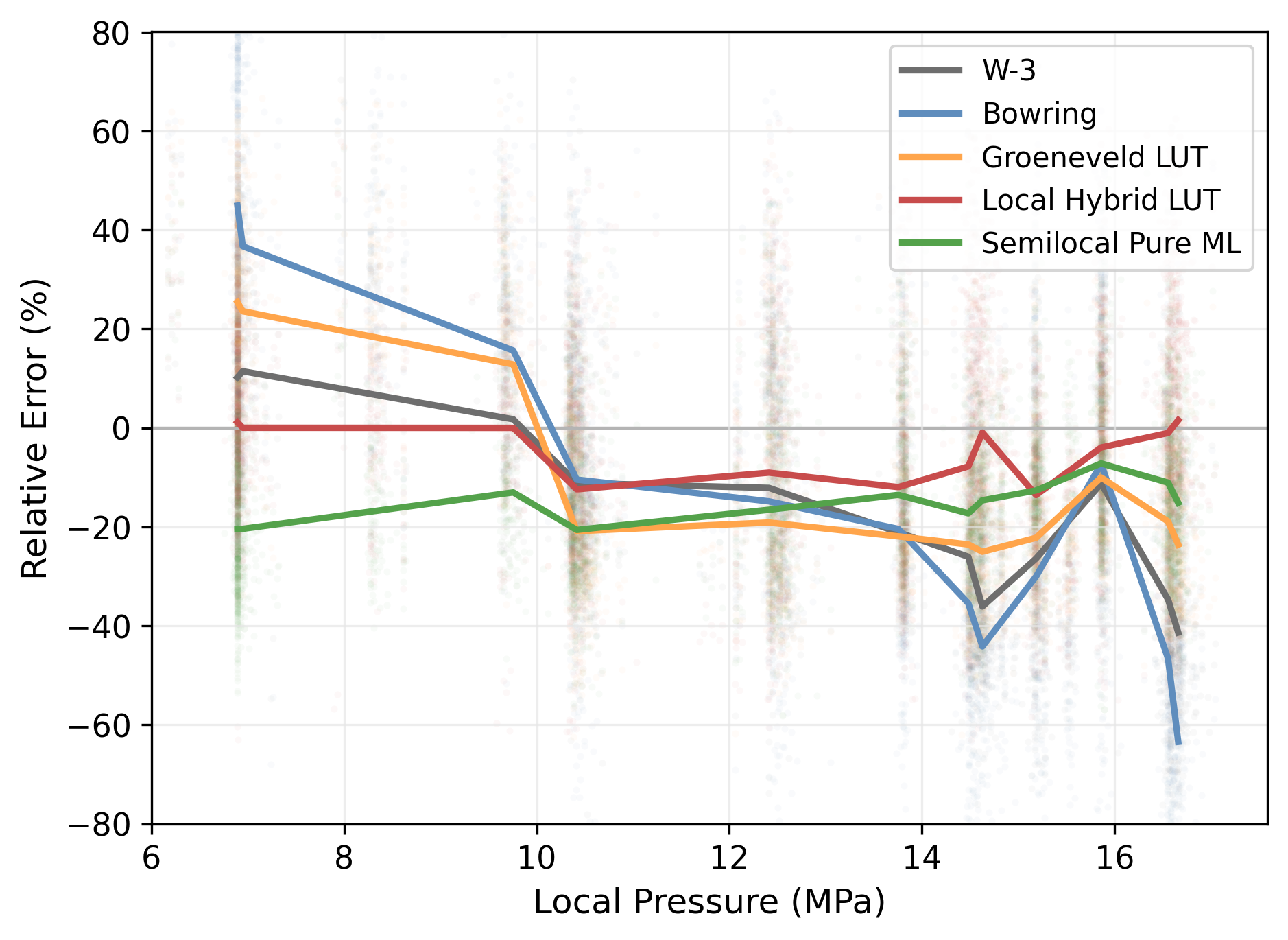}
    \caption{Pressure}
  \end{subfigure}
  \hfill
  \begin{subfigure}{0.49\textwidth}
    \centering
    \includegraphics[width=\linewidth]{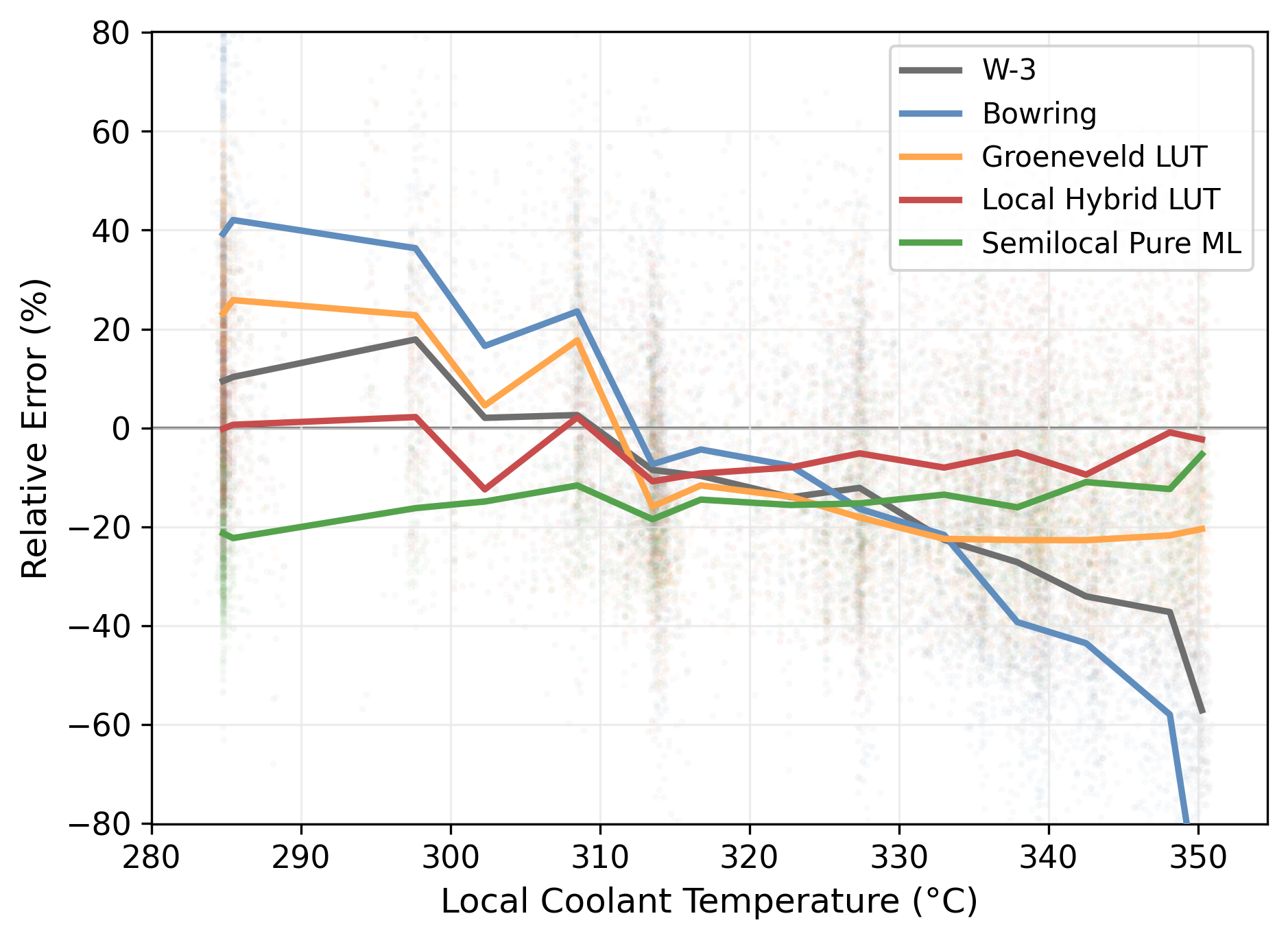}
    \caption{Coolant temperature}
  \end{subfigure}
  \caption{Traditional model baselines compared to the best local (hybrid LUT) and semilocal (pure) ML-based models across local thermal hydraulic conditions at the experimental CHF locations. The curves indicate median relative error values in each method's 12~statistical bins, with the actual points behind.}
  \label{fig:operational_condition_curves}
\end{figure}

Outside of bundle geometry and axial power profile, performance was also examined as a function of the local equilibrium quality, mass flux, pressure, and coolant temperature extracted at the experimental CHF locations. Comparisons between the traditional methods and the best-performing local and semilocal ML-based models are provided in Figure~\ref{fig:operational_condition_curves}. The Bowring correlation is observed to be highly sensitive to all four variables, with error generally increasing at both ends of each variable's range. The W-3 correlation exhibits less sensitivity but still demonstrates a gradual drift towards more negative relative error values at higher mass fluxes, pressures, and temperatures. The 2006 Groeneveld LUT is the most stable of the traditional methods but still sees trends similar to the W-3 correlation with respect to all four variables.

The best-performing local ML-based model, the hybrid LUT, exhibits the most consistent behavior of all evaluated methods. Median errors are shown remaining within 10\% across nearly all bins of equilibrium qualities, mass fluxes, pressures, and coolant temperatures considered in this study. The semilocal pure ML model also demonstrates strong performance, although with a somewhat larger degree of variation and negative bias compared to the local hybrid LUT. Both ML-based models generally avoid the pronounced degradation at the operational ranges' extrema observed in the traditional methods.

Taken together, the bundle geometry, axial power profile, and operational condition analyses indicate that the observed performance improvements are not limited to a specific subset of the evaluation database. While performance differences between models remain dependent on the particular test grouping considered, the ML-based methods generally maintain favorable error characteristics across the range of bundle geometries and operating conditions represented in the EPRI database. The local hybrid LUT model is particularly notable in this regard, consistently producing low and relatively stable errors across nearly all investigated conditions.

%%%%%%%%%%%%%%%%%%%%%%%%%%%%%%%%%%
\section{Conclusions}
\label{sec:conclusions}
%%%%%%%%%%%%%%%%%%%%%%%%%%%%%%%%%%
% STATUS: DRAFTED

This work evaluated CTF’s tube-trained ML-based CHF models when used in square rod bundle simulations from the EPRI database. These models were trained using the NRC public CHF database and evaluated against conventional bundle CHF prediction methods, including the Bowring correlation, Groeneveld LUT, and W-3 correlation. DNN-based pure and hybrid bias correction ML models in both fully local and semilocal formulations were considered.

Prior to model evaluation, the tube training domain was compared to the rod bundle deployment domain, and significant differences were found between the two datasets' distributions. While the rod bundle data mostly remained within the ranges of the tube training data, many rod bundle experiments occupied sparse regions with heated equivalent diameter as the primary source of domain shift. In addition to a simple shift in distribution, the heated equivalent diameter also represents a change in physical interpretation from the tube diameter variable present in the training data.

While these differences are nontrivial, the ML-based CHF models generally demonstrated favorable transferability to rod bundle applications. In most cases, the 2006 Groeneveld LUT was the strongest-performing traditional prediction method, while the Bowring correlation consistently produced the least favorable results. With the exception of the local hybrid Bowring model, the ML-based models broadly improved predictions across error metrics. The local hybrid LUT model, in most geometries and operational conditions, significantly outperformed the other CHF models across error metrics. Consistent error behavior was noted with respect to local equilibrium qualities and mass fluxes, while the traditional models saw stronger region-dependent performance. The semilocal pure ML model broadly obtained the second most favorable error metrics and the most favorable in the case of aggregate $5\times5$ geometry statistics.

Overall, these results indicate that ML-based CHF models originally trained on tube databases can provide improvements in rod bundle CHF prediction even when applied outside of their original training domain. Immediate performance improvements may be possible by selectively retraining tube-based models using a subset of the rod bundle database to establish the identifiability of $D_{\mathrm{he}}$ effects and other bundle-specific behavior. Based on the results, additional bundle-related input features may be considered to attempt to encode physics from sources such as spacer grids and nonuniform power distributions. Quantifying and mitigating uncertainty from non-CHF closure models remains an open challenge to be further investigated.

%%%%%%%%%%%%%%%%%%%%%%%%%%%%%%%%%%%%%%%%%%%%%%%%%%%%%%%%%%%%%%%%%%%%%%%%%%%%%%%%
\section*{Acknowledgments}
%%%%%%%%%%%%%%%%%%%%%%%%%%%%%%%%%%%%%%%%%%%%%%%%%%%%%%%%%%%%%%%%%%%%%%%%%%%%%%%%

The authors from North Carolina State University were funded by the US Department of Energy Office of Nuclear Energy (DOE-NE) Distinguished Early Career Program (DECP) under award number DE-NE0009467. The authors from the University of Wisconsin--Madison were funded by the DOE-NE DECP award DE-NE0009425. Any opinions, findings, and conclusions or recommendations expressed in this paper are those of the authors and do not necessarily reflect the views of the US Department of Energy.

Notice: This manuscript has been authored by UT-Battelle LLC under contract DE-AC05-00OR22725 with the US Department of Energy (DOE). The US government retains and the publisher, by accepting the article for publication, acknowledges that the US government retains a nonexclusive, paid-up, irrevocable, worldwide license to publish or reproduce the published form of this manuscript, or allow others to do so, for US government purposes. DOE will provide public access to these results of federally sponsored research in accordance with the DOE Public Access Plan (https://www.energy.gov/doe-public-access-plan).

\newpage
%%%%%%%%%%%%%%%%%%%%%%%%%%%%%%%%%%%%%%%%%%%%%%%%%%%%%%%%%%%%%%%%%%%%%%%%%%%%%%%%
%%%%%%%%%%%%%%%%%%%%%%%%%%%%%%%%%%%%%%%%%%%%%%%%%%%%%%%%%%%%%%%%%%%%%%%%%%%%%%%%
%%\section*{References}
\bibliography{./bibliography.bib}

\end{document}